%% file: main.tex
\documentclass[10pt,twocolumn,letterpaper]{article}

\usepackage{cvpr}
\usepackage{times}
\usepackage{epsfig}
\usepackage{graphicx}
\usepackage{amsmath}
\usepackage{amssymb}

\usepackage[sort,nocompress]{cite}
\usepackage{xcolor,booktabs,blindtext,subcaption}

\usepackage[pagebackref=false,breaklinks=true,colorlinks,bookmarks=false]{hyperref}

\DeclareGraphicsExtensions{.pdf,.jpg,.png}
\graphicspath{{images/}}

\newcommand{\figref}[1]{Figure~\ref{fig:#1}}

\newcommand{\hquad}{\hspace{0.75em}}
\cvprfinalcopy 

\def\cvprPaperID{61} 

\begin{document}

\title{RasterNet: Modeling Free-Flow Speed using LiDAR and Overhead Imagery}

\author{
    Armin Hadzic$^1$ \hquad Hunter Blanton$^1$ \hquad Weilian Song$^2$ \hquad Mei Chen$^1$ \hquad Scott Workman$^3$ \hquad Nathan Jacobs$^1$\\
    $^1$University of Kentucky \quad $^2$Simon Fraser University \quad  $^3$DZYNE Technologies   
}
\maketitle

\begin{abstract}
  
  Roadway free-flow speed captures the typical vehicle speed in low traffic conditions. Modeling free-flow speed is an important problem in transportation engineering with applications to a variety of design, operation, planning, and policy decisions of highway systems. Unfortunately, collecting large-scale historical traffic speed data is expensive and time consuming. Traditional approaches for estimating free-flow speed use geometric properties of the underlying road segment, such as grade, curvature, lane width, lateral clearance and access point density, but for many roads such features are unavailable. We propose a fully automated approach, RasterNet, for estimating free-flow speed without the need for explicit geometric features. RasterNet is a neural network that fuses large-scale overhead imagery and aerial LiDAR point clouds using a geospatially consistent raster structure. To support training and evaluation, we introduce a novel dataset combining free-flow speeds of road segments, overhead imagery, and LiDAR point clouds across the state of Kentucky. Our method achieves state-of-the-art results on a benchmark dataset.
\end{abstract}


\input{1_intro}
\input{2_related}
\input{3_dataset}
\input{4_method}
\input{5_evaluation}
\input{6_conclusion}

{\small
\bibliographystyle{ieee_fullname}
\bibliography{biblio}
}

\end{document}

%% file: 1_intro.tex
\section{Introduction}

Free-flow speed is defined as the average speed a motorist would travel on a given road segment when it is not impeded by other vehicles. This is an important measure used in transportation engineering for a variety of applications such as traffic control, highway design, measuring travel delay, and setting speed limits. Existing approaches for collecting measurements of free-flow speed have largely been manually intensive and difficult to scale~\cite{deardoff2011estimating}, putting a large strain on transportation engineering budgets. Only recently have more advanced techniques, such as probe vehicles, been used for road performance monitoring~\cite{rouphail2017application}. To avoid the upfront cost of collecting traffic speed data, a variety of recent work has explored developing automatic methods for estimating free-flow speeds.

\begin{figure}
  \centering
  \begin{subfigure}{.5174\linewidth}
    \includegraphics[width=\linewidth]{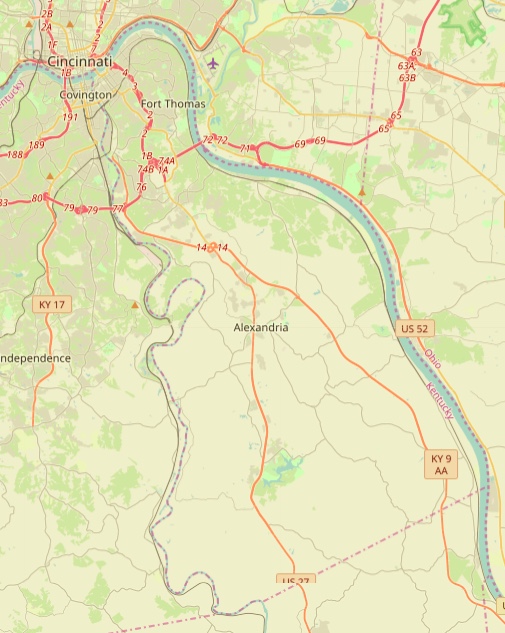}
  \end{subfigure}
  \begin{subfigure}{.4626\linewidth}
    \includegraphics[width=1\linewidth]{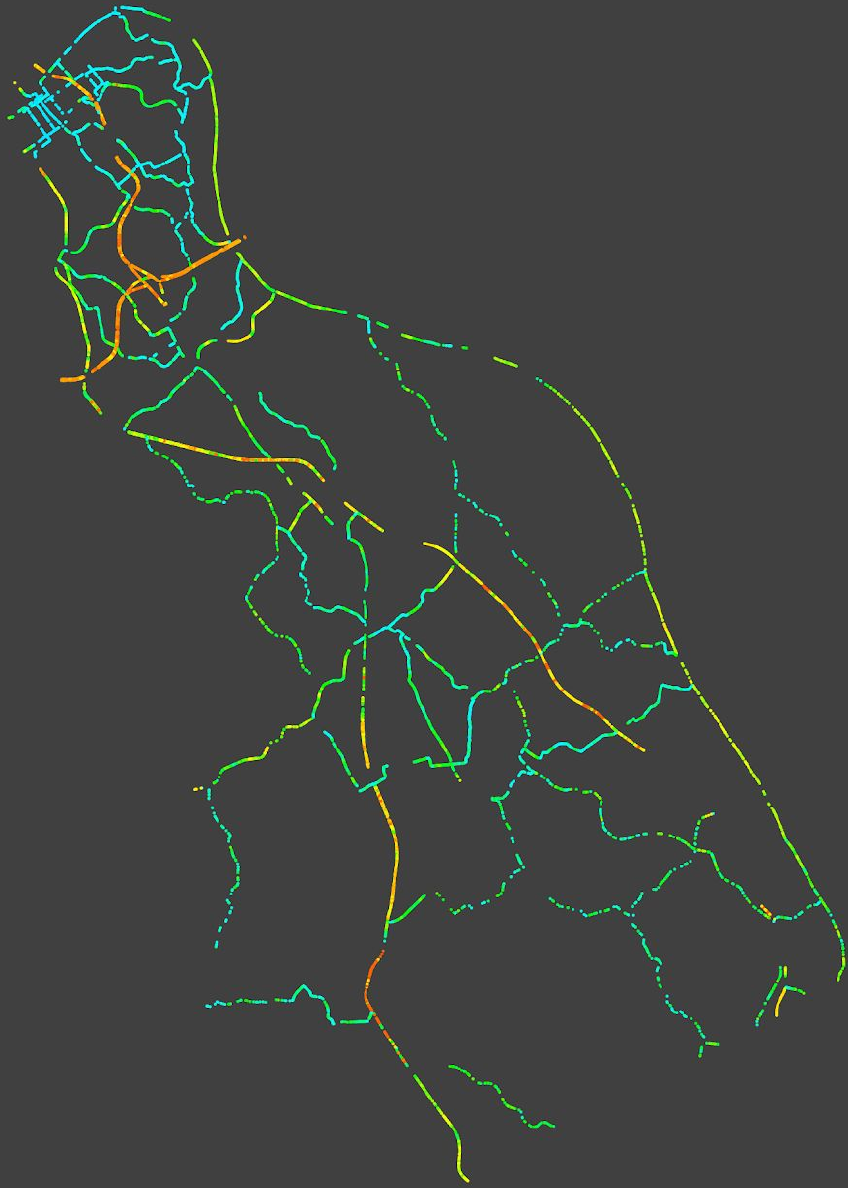}
  \end{subfigure}
    \caption{We propose an automatic approach for estimating free-flow speed from overhead imagery and 3D airborne LiDAR data. (left) A map representing Campbell county in Kentucky, USA. (right) The corresponding map of free-flow speeds generated using our method.}
  \label{fig:cartoon}
\end{figure}

Traditional approaches for free-flow speed modeling involve the use of geometric road features (also known as highway geometric features) such as lane width, lateral clearance, median type, and access points~\cite{manual2010hcm2010}. These approaches tend to be specific to certain road network types (arterial, local, collector)~\cite{silvano2016impact}, or geographical areas (urban and rural)~\cite{sekhar2016free}. While these methods have demonstrated good performance, their use is limited to areas where the necessary road metadata is available. Typically, these areas include state-maintained highways such as interstates, US highways, and state roads. However, this is often a small portion of all roads. For example, only 35\% of all roadway miles in Kentucky are state-maintained. The detailed geometric features required for estimating free-flow speed on locally maintained roads are mostly unavailable or prohibitively expensive to collect. Estimating free-flow speeds at large scales requires learning-based methods that take advantage of alternative data sources (\figref{cartoon}).

Recent work has shown that road geometry approaches can be augmented with visual data, in the form of overhead imagery, to improve performance~\cite{song2019remote}. Though adding visual features results in better performance than road geometric features alone, model applicability is still limited to sufficiently documented roads. Instead, we explore replacing explicit road geometric features with features extracted from airborne LiDAR (Light Detection and Ranging) point clouds. Compared to image data which is often impacted by transient effects (e.g., weather), 3D point clouds are viewpoint invariant, robust to weather and lighting conditions, and provide explicit 3D information not present in 2D imagery, offering a supplementary source of data. Our approach combines both sources, visual features extracted from overhead imagery and geometric features extracted from point clouds.

We propose \emph{RasterNet}, a multi-modal neural network architecture that combines overhead imagery and airborne LiDAR point clouds for the task of free-flow speed estimation. To align the input domains, \emph{RasterNet} organizes local point cloud neighborhoods using a raster center grid and pairs them with spatially consistent features extracted from the image data. Features from both domains are then merged together and used to jointly estimate free-flow speed. To support the training and evaluation of our methods, we introduce a large  dataset containing free-flow traffic speeds, overhead imagery, and airborne LiDAR data across the state of Kentucky. We evaluate our method both qualitatively and quantitatively, achieving state-of-the-art results compared to existing methods, without requiring explicit geometric features as input. 

Our primary contributions can be summarized as follows:
\begin{itemize}
    \item A large dataset for free-flow speed estimation that combines speed data, overhead imagery, and corresponding point clouds.
    \item A novel multi-modal neural network architecture for free-flow speed estimation that advances the state-of-the-art on an existing benchmark dataset.
    \item A method for fusing overhead imagery and airborne LiDAR point clouds using a geospatially consistent raster structure.

\end{itemize}

%% file: 2_related.tex
\section{Related Work}

We provide an overview of work in three related fields: point cloud representations multi-modal data fusion, and estimating traffic speed.


\subsection{Point Cloud Representations}

Many methods have been proposed for extracting feature representations from point clouds. Recently, Weinmann et al.~\cite{weinmann2014semantic}, Liu et al.~\cite{Liu2018LPDNet3P}, and Dub\'e et al.~\cite{segmatch2017} demonstrated that point clouds could be represented by neighborhood structural statistics in order to improve performance on scene understanding and place recognition tasks. The seminal work of Qi et al.~\cite{pointnet} introduced PointNet, a general deep neural network for point cloud feature extraction. This work inspired a series of works in point cloud shape classification~\cite{thomas2019KPConv, mao2019interpolated, lan2019modeling} and object detection~\cite{shi2019pointrcnn}. Later, Qi et al. presented PointNet++~\cite{pointnet++}, a shape classification method and extension to PointNet which adds local feature extraction to improve performance. This method allows for precise control over the spatial location of extracted features, which we use for geospatially aligning point cloud features with visual features from an image.




\subsection{Multi-Modal Data Fusion}

A significant amount of work has explored combining imagery with LiDAR data for various tasks. Liang et al.~\cite{liang2018deep} designed a method for multi-scale fusion of ground imagery with overhead LiDAR point clouds to perform object detection from multiple viewpoints and modalities. Similar to our own work, Jaritz et al.~\cite{jaritz2019xmuda} used a cross-modal autonomous driving dataset to perform unsupervised domain adaption for 3D semantic Segmentation. Their dataset combined terrestrial LiDAR point clouds and camera images for different times of day, countries, and sensor setups. Their proposed  cross-modal model, xMUDA, performs data fusion by projecting 3D point cloud points onto the 2D image plane and sampling features at corresponding pixel locations. While this dataset and method were designed for small spatial areas around a vehicle, we perform data fusion of overhead imagery and airborne LiDAR point clouds of large $200 \times 200m^2$ areas. 

Recent work has also explored the fusion of airborne LiDAR with overhead imagery for the task of semantic segmentation in an urban area~\cite{daneshtalab2019cnn}. Typically these approaches render the LiDAR data as 2D images through digital surface models and use a traditional CNN. This strategy results in a loss of precise 3D information due to discretization. This is an issue, as raw point cloud methods have been shown to outperform discretization-based approaches for classification tasks~\cite{Mao_2019_ICCV}. Our approach uses point cloud understanding to process 3D point clouds.


\subsection{Estimating Traffic Speed}

Several works have proposed automatic methods for estimating the speed of vehicles. Huang~\cite{huang2018traffic} used video surveillance data of traffic to perform individual vehicle speed estimation. We perform average free-flow speed estimation to characterize traffic flow behavior and capacity of roads instead of individual vehicle speed characteristics. Most similar to our own work, Song et al.~\cite{song2019remote} performed free-flow speed estimation using overhead imagery and geometric road features on the Kentucky free-flow speed dataset. Our \emph{RasterNet} model is trained on the same overhead imagery and label data, but our approach replaces the provided geometric road features with point cloud features of the same spatial area.

%% file: 3_dataset.tex
\begin{figure*}
    \centering
    
    \setlength\tabcolsep{1pt}

    \begin{tabular}{cccc}
        \includegraphics[height=3.2cm]{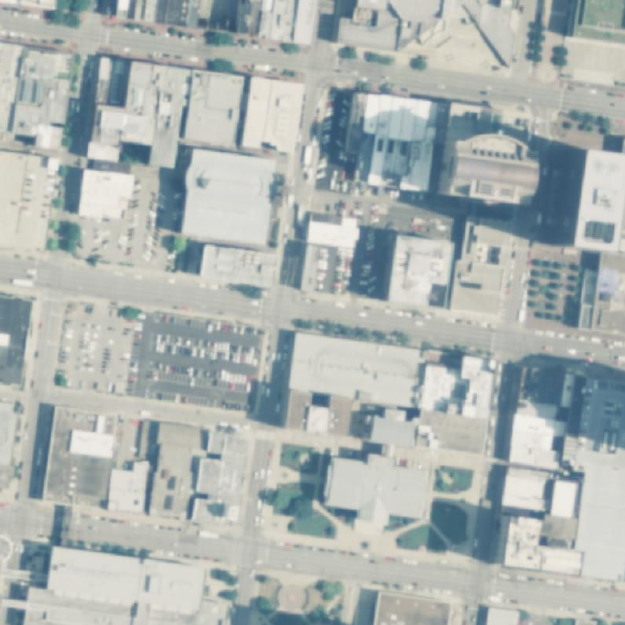} &
        \includegraphics[height=3.2cm,trim=27px 7px 156px 20px, clip]{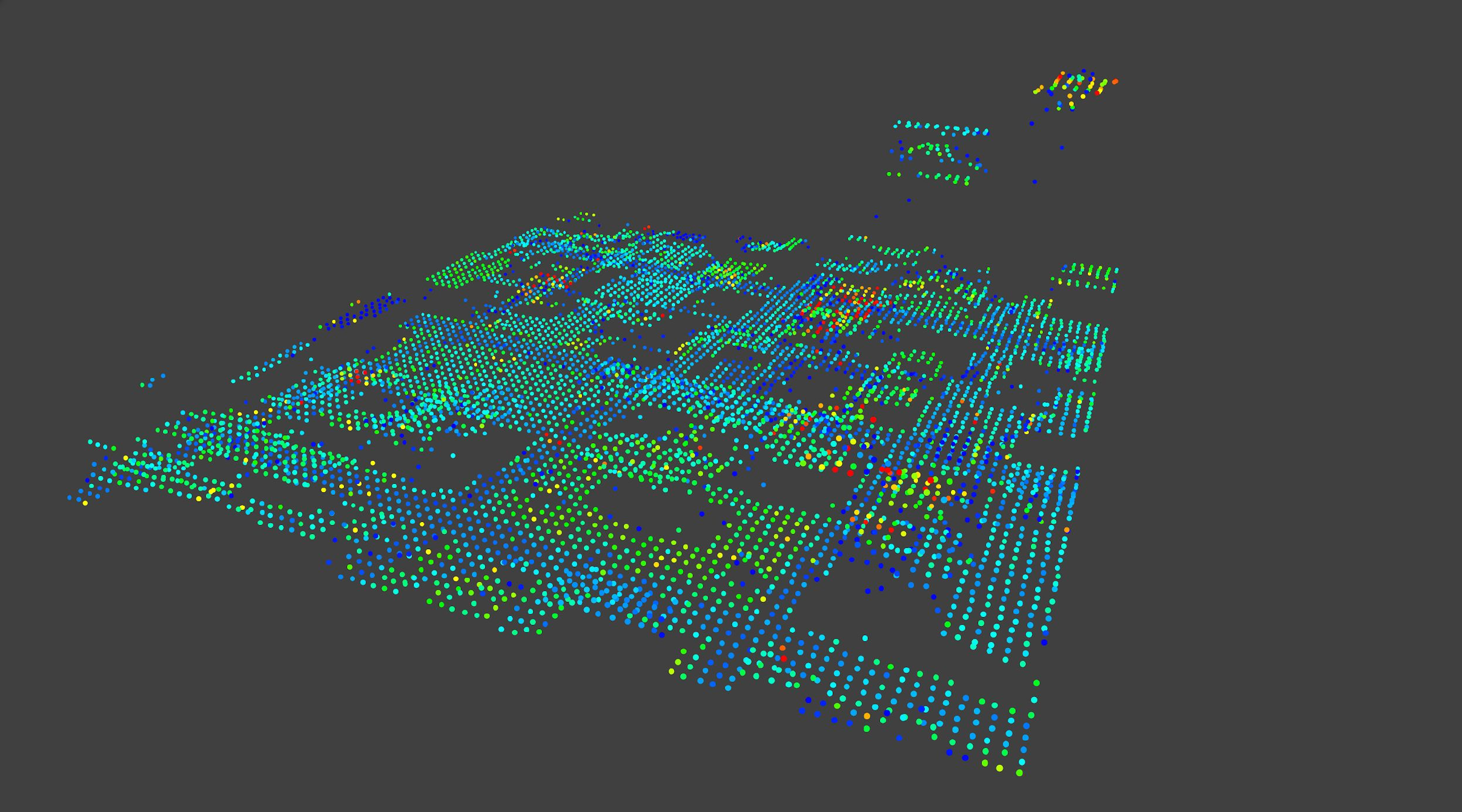} &
        \includegraphics[height=3.2cm]{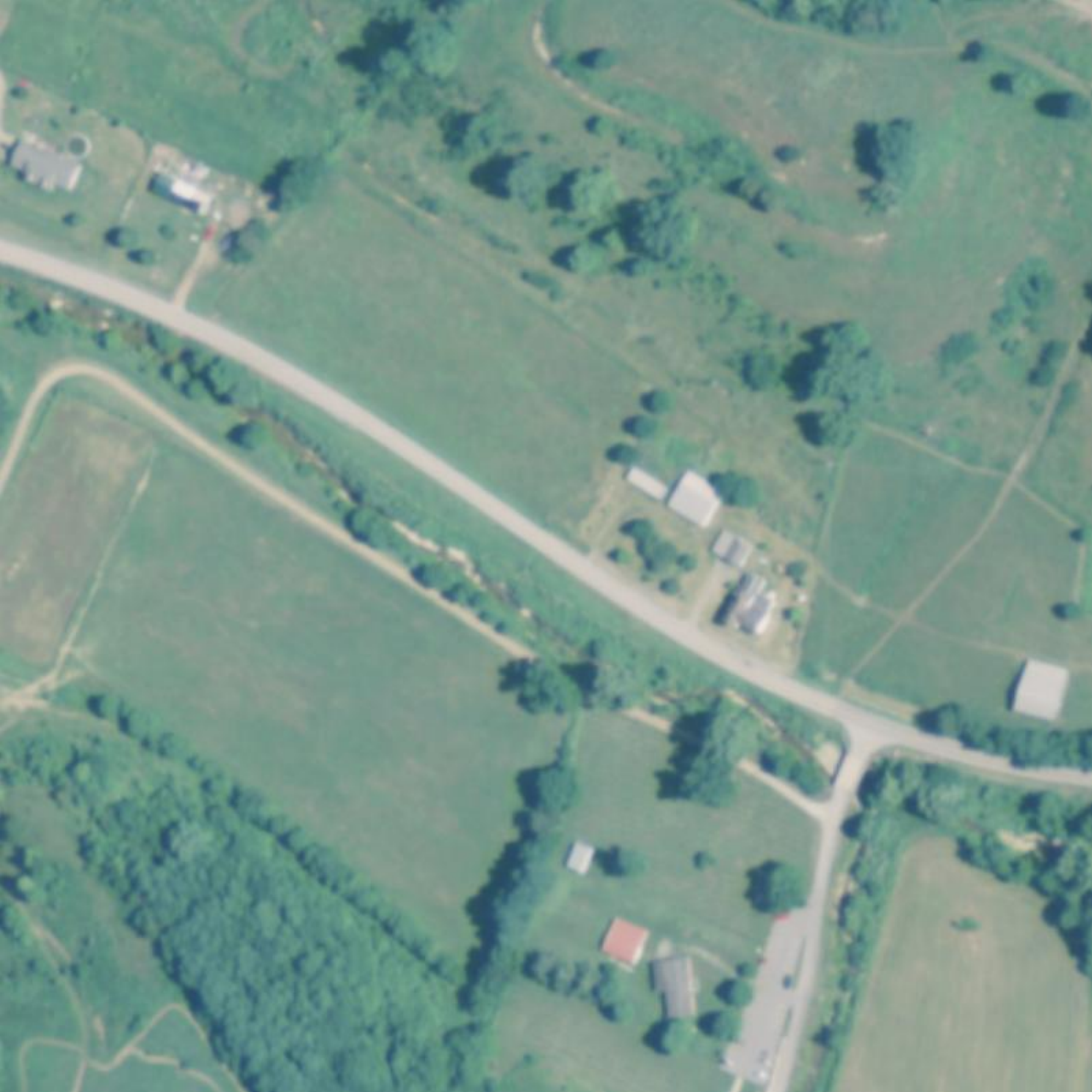} &
        \includegraphics[height=3.2cm,trim=0px 2px 30px 40px, clip]{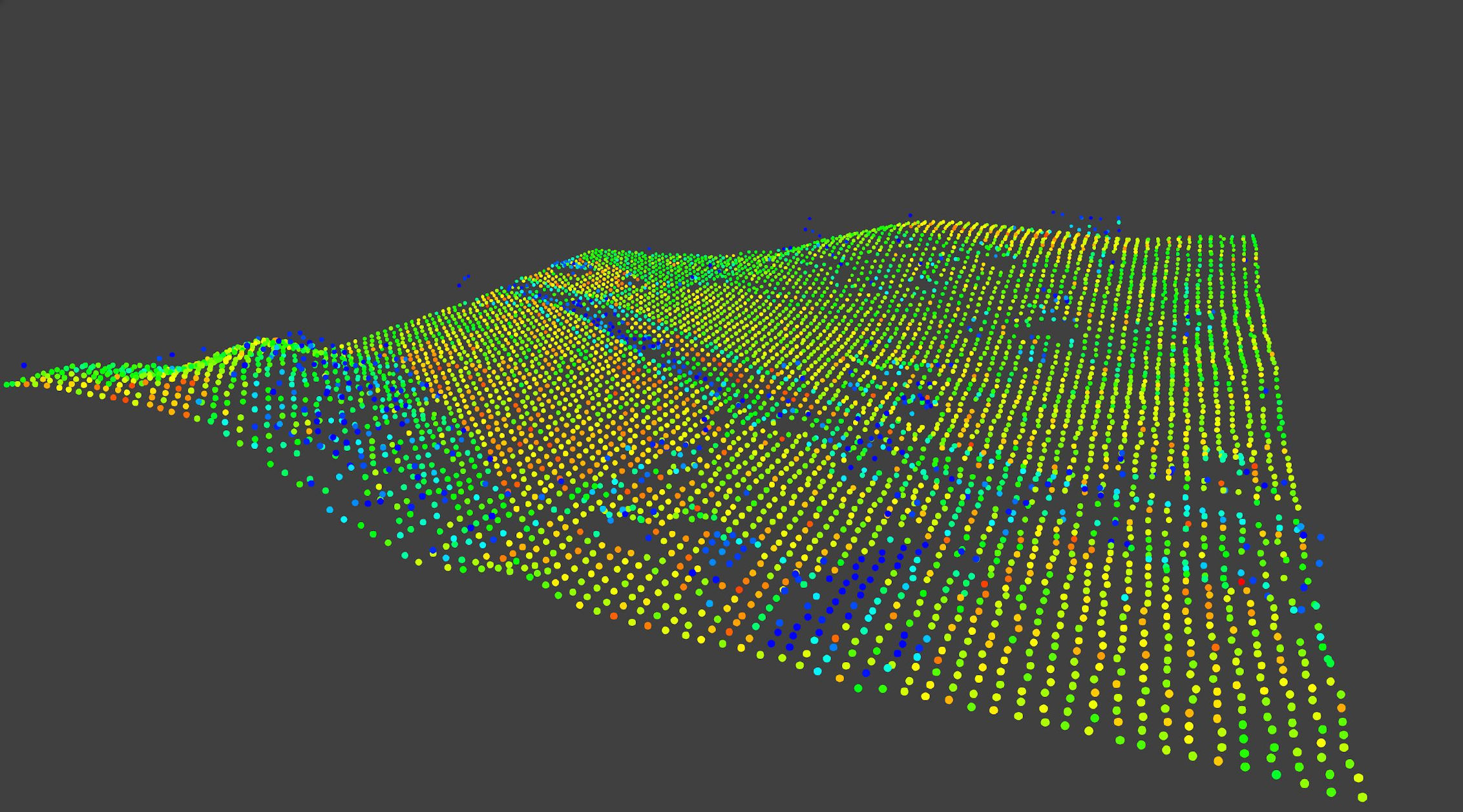} \\
    \end{tabular}
    
    \caption{Examples of our multi-modal dataset. A geospatially aligned overhead image and corresponding point cloud are shown for an urban scene (left) and a rural scene (right). Point cloud coloring represents the relative grayscale intensity.}
    \label{fig:dataset_ex}
\end{figure*}

%
%
%

\section{A Multi-Modal Dataset for Free-Flow Speed Estimation}

We introduce a large-scale dataset for free-flow speed estimation that combines free-flow speed data, point clouds obtained from airborne LiDAR, and overhead imagery. Our dataset extends a recently introduced dataset that relates speed data on road segments throughout Kentucky, USA with overhead imagery. We begin by giving an overview of this existing dataset, then describe how we augment it with geospatially consistent 3D point cloud data.

\subsection{Kentucky Free-Flow Speed Dataset}

The Kentucky Transportation Center~\cite{chen2015analysis} licensed and aggregated HERE Technologies's speed data across uncongested periods to produce free-flow speeds for road segments across Kentucky. The speed data was then spatially joined with the Kentucky Transportation Cabinet's highway inventory data. For each road segment, Song et al.~\cite{song2019remote} collected an overhead image centered at the location of the free-flow speed label. The overhead imagery is from the National Agriculture Imagery Program (NAIP) with 1m ground-sample distance (GSD). A single image has a spatial coverage of $200 \times 200m^2$. Each image was resized to $224 \times 224$ pixels and rotated to ensure the road segment was aligned with direction of travel to the North. The dataset is representative of rural, urban, highway and arterial roads ranging in structure from multi-lane paved roads to single-lane dirt/gravel roads.

\subsection{Augmenting with Point Cloud Data}

We augment this dataset with 3D point clouds extracted from LiDAR data collected by the Kentucky Division of Geographic Information's KyFromAbove~\cite{kyfromabove} program. Unlike overhead images, geometric features such as change in elevation, road curvature, lane delineation markings, lane width, proximity to neighboring structures, and more, can be easily detected from airborne LiDAR point clouds. The LiDAR data was stored as a collection of $1524 \times 1524m^2$ tiles covering the state of Kentucky. To relate point cloud data with geospatially aligned overhead imagery and free-flow speed data, we performed a two-step process consisting of LiDAR tile selection and point cloud sampling. 

In order to associate each free-flow speed label with its containing tile, we constructed an R-tree using each tile's geospatial coordinates. Then for each tile, we constructed a k-d tree over a random subset of points (50\% selected uniformly at random) to support faster nearest-neighbor lookup. To generate an aligned point cloud, we use an $80 \times 80$ uniformly sampled grid to guide point subsampling in a bounding box of the same spatial dimension as the overhead image. The resulting point cloud is centered on the target label location and is used to represent the spatial features of a given road segment.

An overview of our dataset is shown in Figure~\ref{fig:dataset_ex}. The urban road segment depicted in \figref{dataset_ex} (left) corresponds to the point cloud of the same road segment. The point cloud intensities are illustrated by dark blue roadways in stark contrast with the red roof tops of sky scrapers (top right). Similarly, the rural road segment point cloud in \figref{dataset_ex} (right) shows the dynamic topography of the surrounding landscape not present in the corresponding overhead image. 

%% file: 4_method.tex
\begin{figure*}
  \centering
  \includegraphics[width=.9\linewidth]{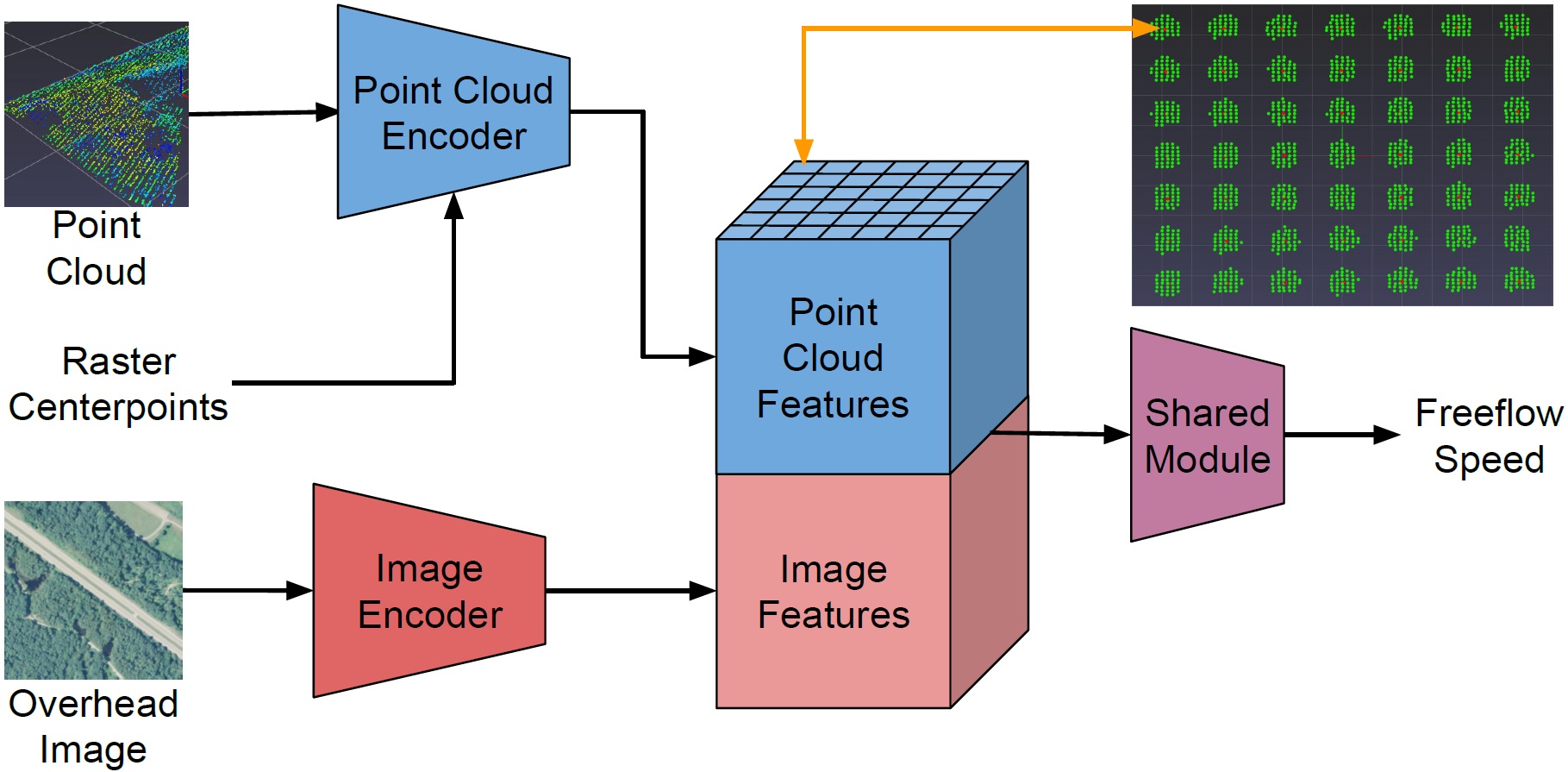}
  \caption{An overview of the \emph{RasterNet} architecture. Overhead images pass through an image encoder, while point clouds and raster center locations are passed through a point cloud encoder. Each cell of the point cloud feature map corresponds to a set of features of a local point cloud neighborhood. The two sets of features are channel-wise concatenated before being passed through a shared model (ResNet block) to produce a free-flow speed prediction.} 
  \label{fig:rasternet_cartoon}
\end{figure*}
\section{Methods}

We introduce \emph{RasterNet}, an architecture for free-flow speed estimation that fuses multi-modal sensory input from overhead images and 3D point clouds. A visual overview of our architecture is given in \figref{rasternet_cartoon}. Overhead images pass through an image encoder, while point clouds and raster center locations are passed through a point cloud encoder. A set of raster center locations guide point cloud feature extraction to produce geospatially consistent features between the two domains. The two sets of features are then channel-wise concatenated before being passed through a shared model to produce a free-flow speed prediction. We describe each component of our architecture in detail in the following sections.

\subsection{Learning Visual Features}
\emph{RasterNet}'s image encoder is based on ResNet~\cite{resnet}, a popular neural network architecture that contains residual connections. Specifically, we chose ResNet18 for our image feature extractor due to its low parameter count and relatively high performance on other tasks such as ImageNet~\cite{imagenet} classification. In this work, we truncate before the average pooling layer such that the final encoding is size $C \times H \times W$, where $C$ refers to the channel dimension and $H$ and $W$ refer to the spatial dimensions of the output feature map. 

\subsection{Extracting Point Cloud Features}

We explore two strategies for extracting point cloud features: (1) using a learning-based method (\emph{RasterNet Learn}), and (2) using features computed from structural statistics (\emph{RasterNet Statistics}). We begin by describing how we define a grid of point locations to align point cloud features with visual features.

\subsubsection{Aligning Visual and Point Cloud Features}

\begin{figure}
  \centering
  \includegraphics[trim=5px 0px 5px 0px, clip, width=\linewidth]{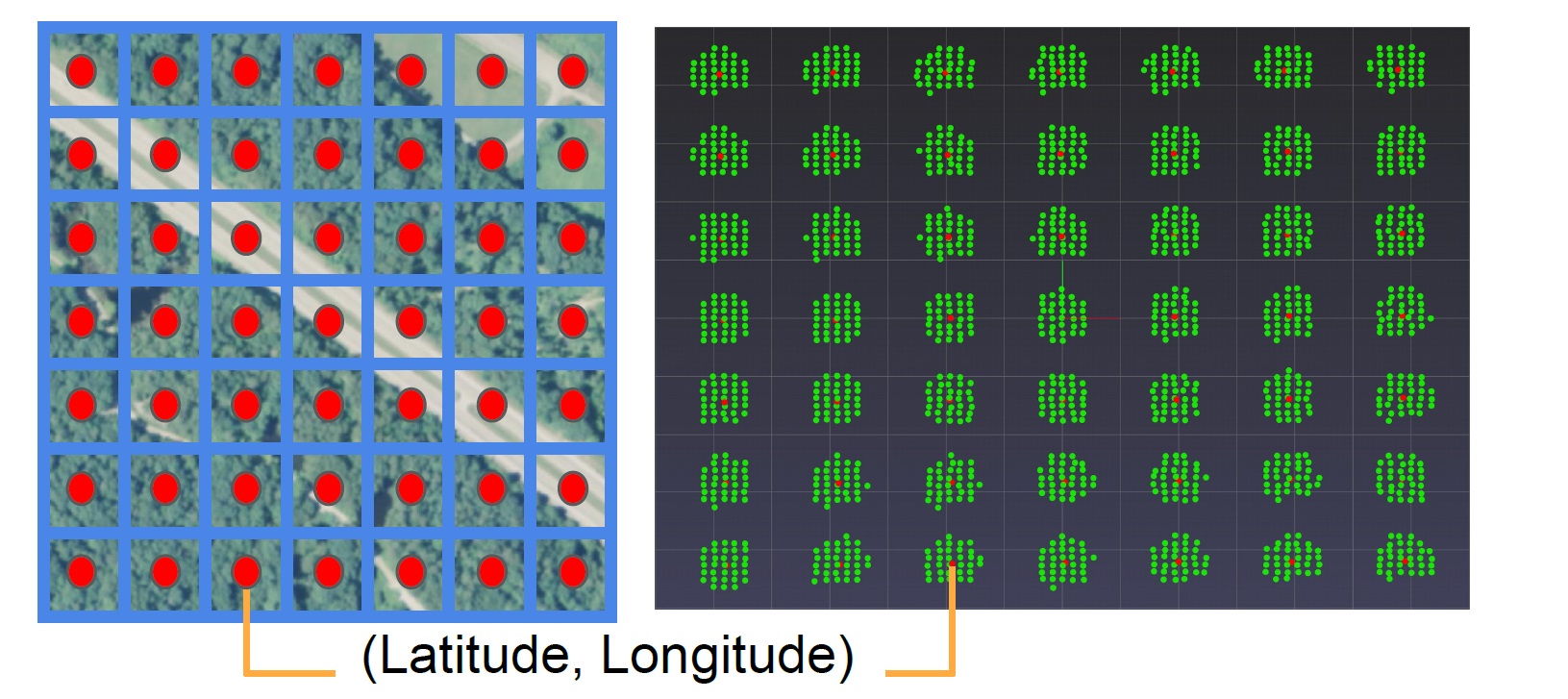}
  \caption{Image features are paired with point cloud features using a grid of raster center points (red dots), ensuring geospatial consistency between the two feature sets.} 
  \label{fig:raster_pairing}
\end{figure}

An inherent challenge of training deep learning models on point clouds is their lack of fixed and consistent structure. To guide point cloud feature extraction, we propose a structural tool, the raster center grid, to impose consistent structure on extracted point cloud features. As \figref{raster_pairing} illustrates, each raster center (red dots) binds local neighborhoods of point cloud features to a fixed location in a $H \times W$ grid, similar to how CNNs group image features. The raster center grid was constructed to geospatially align with the pixel locations of the 2D image encoding. We did this by linearly sampling an $H \times W$ grid within the known bounding box of the overhead image. This enables features extracted from point clouds to be directly paired with image features in a geospatially consistent manner.

\subsubsection{Learned Features}

The \emph{RasterNet Learn} model uses a modified PointNet++~\cite{pointnet++} architecture as a learned point cloud feature extractor. PointNet++ was selected as a point cloud feature extractor because of its simplicity and high accuracy on point cloud tasks. The publicly available PyTorch~\cite{pytorch} implementation of PointNet++ from Wijmans~\cite{pytorchpointnet++} was modified so the second multi-scale grouping layer performed grouping around the raster center grid of a given point cloud instead of using furthest point sampling. This modification allows the point cloud features to be combined with image features while maintaining spatial consistency. After the second multi-scale grouping layer the remainder of Pointnet++ was replaced with a series of $1 \times 1$ convolutions that reduced the number of collected features per raster center to 16.

\begin{figure}
    \centering
    \begin{subfigure}{.49\linewidth}
        \includegraphics[width=\linewidth]{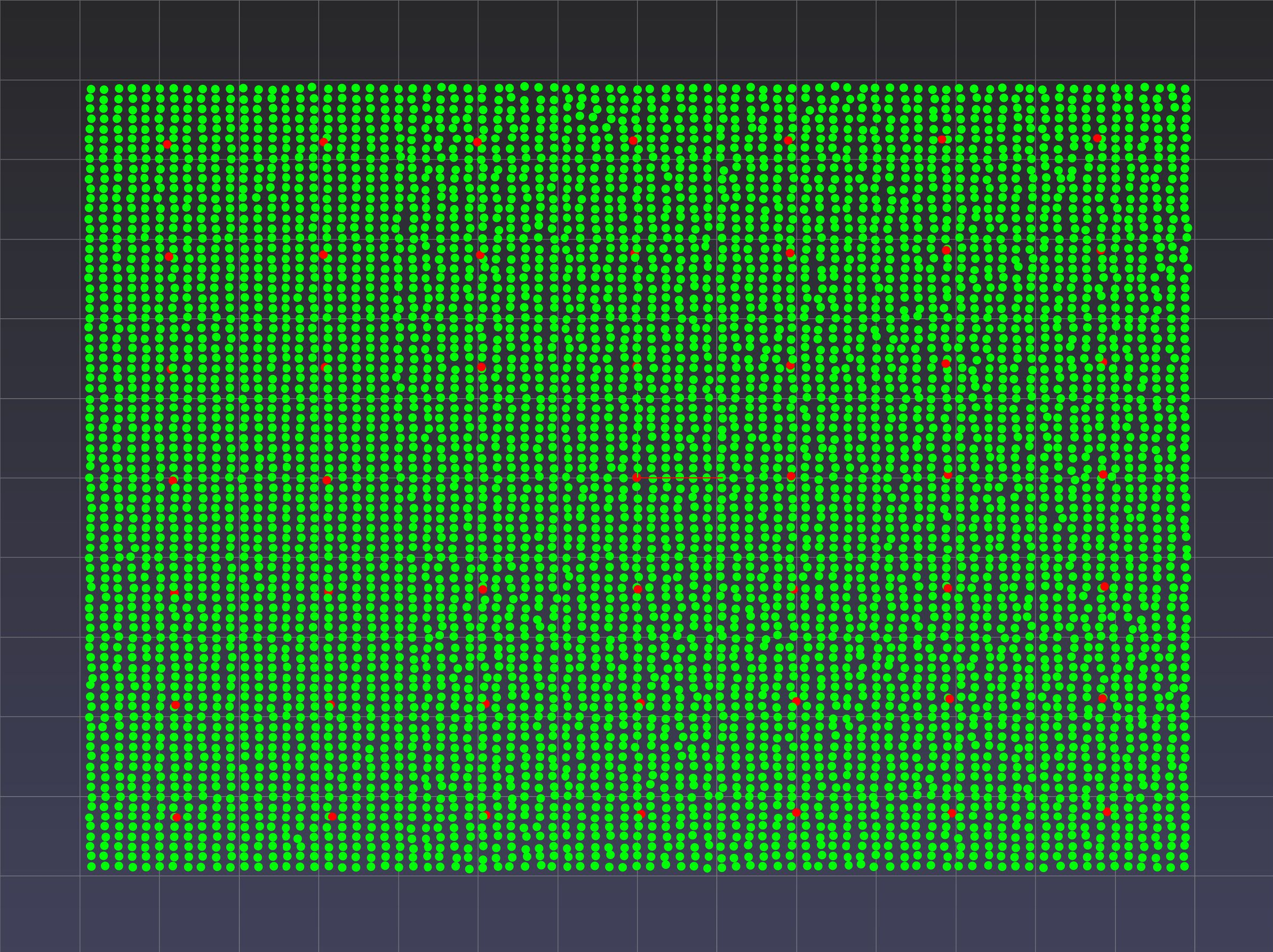}
        \caption{Full Point Cloud}
    \end{subfigure}
    \begin{subfigure}{.49\linewidth}
        \includegraphics[width=\linewidth]{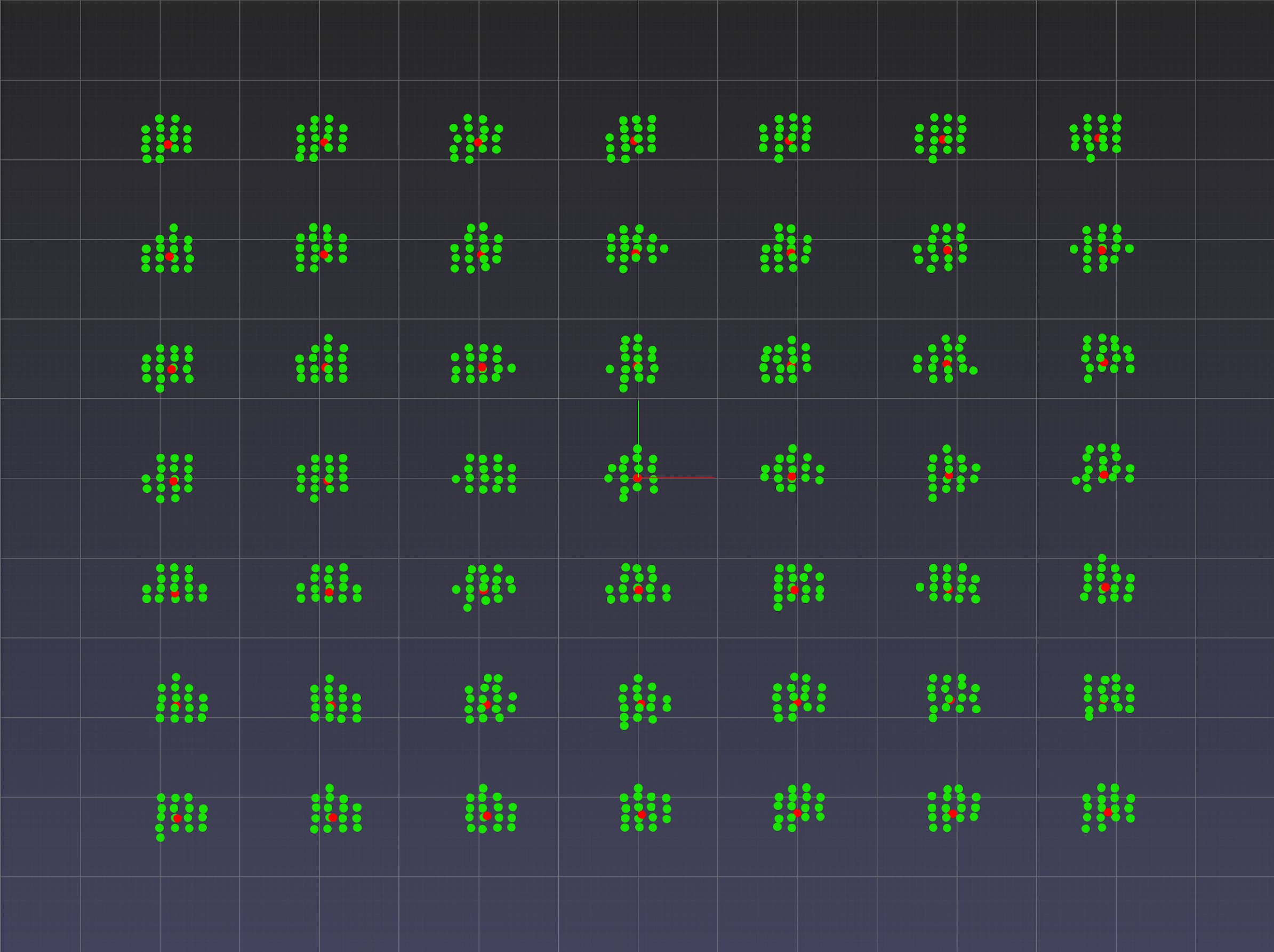}
        \caption{Grouping 16 Samples}
    \end{subfigure}
    \begin{subfigure}{.49\linewidth}
        \includegraphics[width=\linewidth]{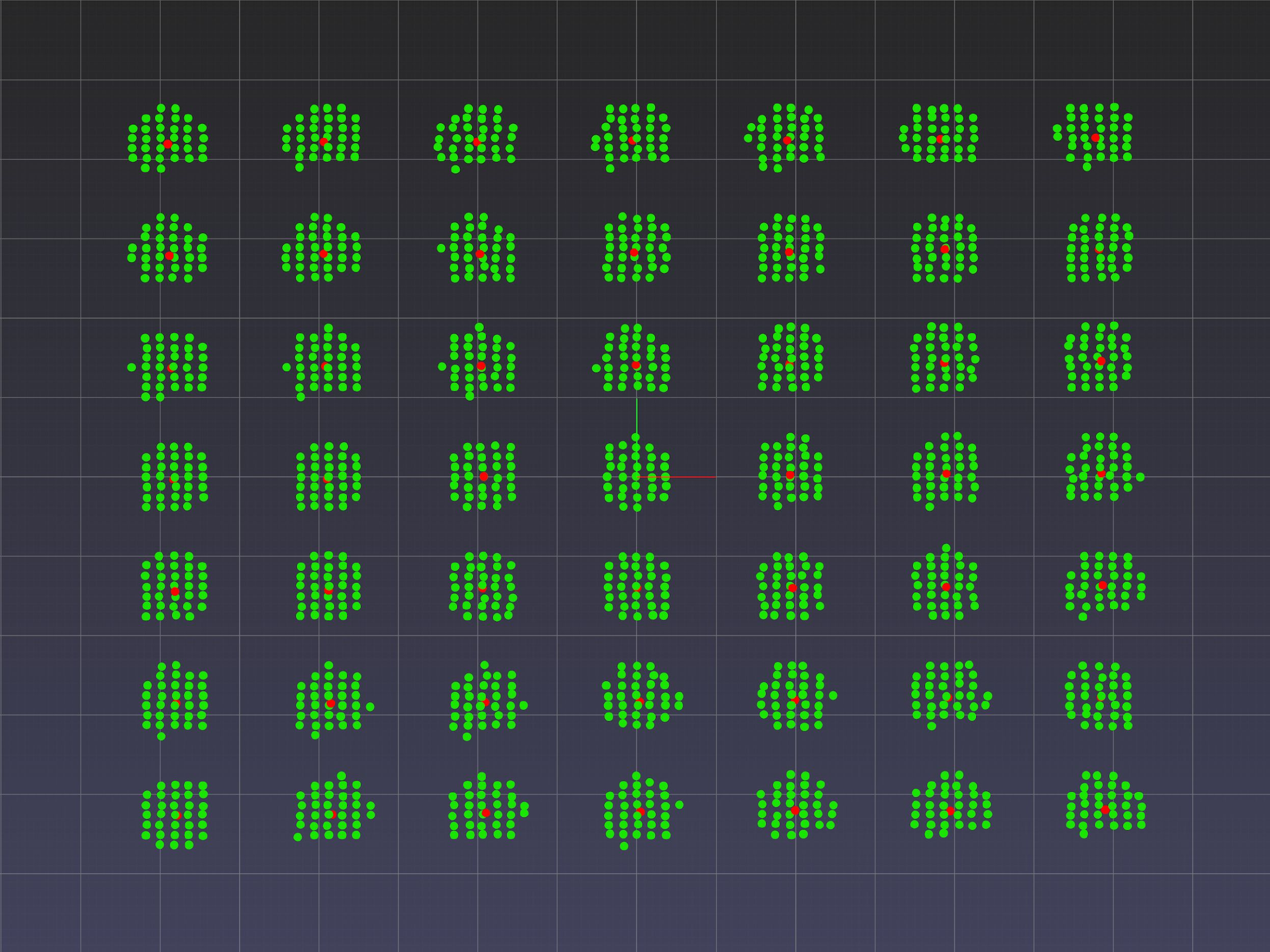}
        \caption{Grouping 32 Samples}
    \end{subfigure}
    \begin{subfigure}{.49\linewidth}
        \includegraphics[width=\linewidth]{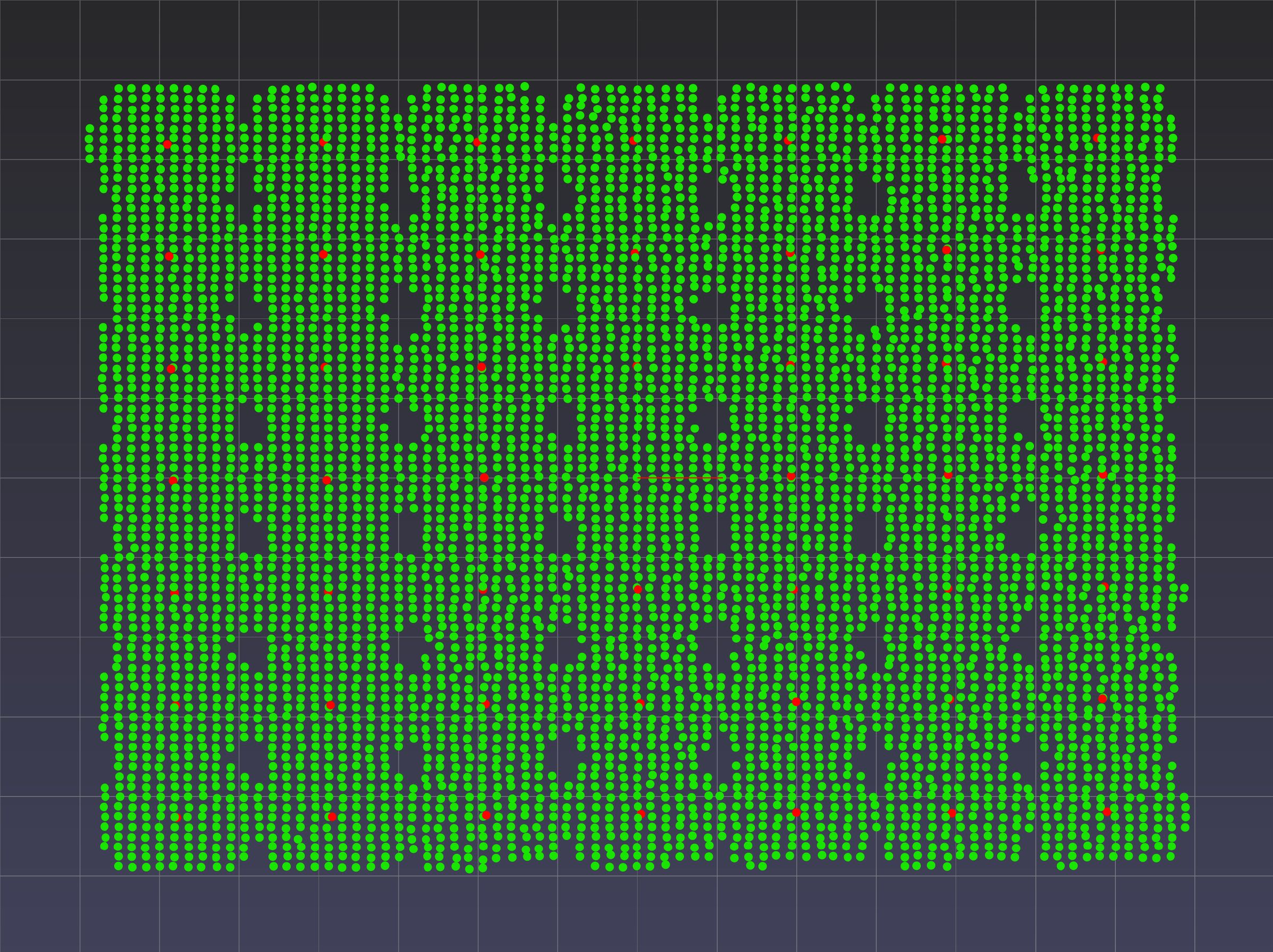}
        \caption{Grouping 128 Samples}
    \end{subfigure}
    
    \caption{PointNet++~\cite{pointnet++} style multi-scale grouping depicted for a point cloud (a) centered on a known free-flow speed label location. Grouping operations are performed around each of the raster centers (red) at different scales and number of samples. Local point clouds (green) are grouped at varying sample sizes: 16 samples (b), 32 samples (c), and 128 samples (d).}
    \label{fig:pnpp_grouping}
\end{figure}


\begin{table}
  \renewcommand{\arraystretch}{2}
  \centering
  \caption{Structural Statistics}
  \begin{tabular}{@{}ll@{}}
    \toprule
    \multicolumn{1}{l}{Structural Statistic} & \multicolumn{1}{l}{Equation} \\
    \hline
    {\em Change of Curvature}                       & $C = \frac{\lambda_3}{\lambda_1 + \lambda_2 + \lambda_3}$ \\
    {\em Omni-variance}                             & $O = \frac{\sqrt[3]{\lambda_3}}{\lambda_1 + \lambda_2 + \lambda_3}$ \\
    {\em Linearity}                                 & $L = \frac{\lambda_1 - \lambda_2}{\lambda_1}$ \\
    {\em Eigenentropy}                              & $A = -\sum_{j=1}^{3} \lambda_j \ln{\lambda_j}$ \\
    {\em Local Point Density}                       & $D = \frac{k}{\frac{4}{3} \prod_{j=1}^{3} \lambda_j}$ \\
    {\em 2D Scattering}                             & $S_{2D} = \lambda_1^{2D} + \lambda_2^{2D}$ \\
    {\em 2D Linearity}                              & $L_{2D} = \frac{\lambda_2^{2D}}{\lambda_1^{2D}}$ \\
    {\em Verticality}                               & $V = v_{3,z}$ \\
    {\em Max Height Difference}                     & $\Delta Z = max(x_z) - min(x_z)$ \\
    {\em Height Variance}                           & $\sigma^2 = \frac{1}{N}\sum\limits_{i=1}^N( x_{z_i} - \overline{x_z})^2 $ \\
    \bottomrule
  \end{tabular}
  \label{tbl:struct_eqs}
\end{table}

\subsubsection{Statistical Features}

Alternatively, we also developed the \emph{RasterNet Statistics} model that directly extracts structural statistic features from the input point cloud. The \emph{RasterNet Statistics} model replaced the PointNet++ architecture from \emph{RasterNet Learn} model with a single instance of multi-scale grouping, as depicted in Figure~\ref{fig:pnpp_grouping}. This approach allowed the model to aggregate spatial features at small, medium, and large scales. A single-scale grouping operation collects groups of points around each of the raster center points. Multi-scale grouping transformed each input point cloud into three separate collections point clouds, each for a different neighborhood group size $k$.

Inspired by Liu et al's~\cite{Liu2018LPDNet3P} work on place recognition using LiDAR point cloud structural features, we extracted statistical features from airborne LiDAR point clouds. Let $x_{i,k}$ be a point cloud containing $k$ neighborhood points around point $i$. Neighborhood statistical features were extracted by first calculating the covariance matrix of $x_{i,k}$. We compute eigenvalues of the covariance matrix, resulting in three eigenvalues $\lambda_1 \ge \lambda_2 \ge \lambda_3 \ge 0$. The structural statistics of the point cloud were calculated according to the equations listed in Table~\ref{tbl:struct_eqs}. Note, 2D statistics (Scattering and Linearity) were calculated using 2D eigenvalues, which were calculated from the covariance matrix of the 3D point cloud projected to the xy-plane. We use $x_z$ to specify that we consider only the $z$ component of the points in point cloud $x$, and $x_{z_i}$ to express the $i$th point in $x_z$. For verticality~\cite{weinmann2014semantic}, $v_{3,z}$ is the $z$ component of the eigenvector corresponding to the smallest eigenvalue, $\lambda_3$.

In order to get statistical features at local point cloud regions, we extracted 10 statistical features for each of the local point cloud neighborhoods corresponding to each raster center. We compute this for three different group sizes $k \in \{16, 32, 128\}$, resulting in 30 total features per local point cloud neighborhood. The structure features of each raster center are tiled to create a single $30 \times H \times W$ feature map, corresponding to the feature map resolution of the image encoding. This is then reduced to $10 \times H \times W$ by applying a series of $1 \times 1$ convolutions.  




\subsection{Feature Fusion for Estimating Free-Flow Speed}

The visual features and point cloud features are channel-wise concatenated and passed through a shared module whose role is to extract high-level features from the combined domains and produce a free-flow speed estimate. The spatial correspondence established by the raster center grid between the overhead image and point cloud features ensured that the two sets of input features are spatially aligned. To represent the shared module, we use a single ResNet18 block and a drop out layer for regularization followed by a fully connected layer with $K$ outputs. 

\subsection{Implementation Details}

We model the free-flow speed prediction as a multi-class classification problem. Free-flow speeds were binned into $K=79$ possible classes, each in 1mph increments. Our models train using the cross-entropy loss ($L$) with a \emph{softmax} activation defined as follows,
\begin{equation}
    L(Y, \hat{Y})=-\frac{1}{N}\sum^{N}_{i=  1} \log\left(\frac{e^{y_i}}{\sum^{K}_{j}e^{\hat{y}_{i,j}}}\right).
    \label{eq:crossentropy}
\end{equation}
Let $y_i \in Y$ be a positive class bin label for the $i$th sample from $N$ training samples. The predicted probability from the distribution $\hat{Y}$ for the $i$th sample from the $j$th class was expressed as $\hat{y}_{i,j}$, where $j \in \{0,1,\cdots,K\}$.

The x and y dimensions of point clouds in dataset were translated such that the origin corresponds to the center of the matching overhead image. The height dimension of the point cloud was normalized by subtracting from the median height for the given point cloud and the point intensity values were normalized by dividing by 255. All point clouds were then rotated such that the direction of travel of the target road was pointed north, similar to how the imagery was aligned.

Given an input image of size $224 \times 224$, the output feature map of the image encoder is of size $C \times 7 \times 7$. Therefore, we define the raster center grid to be of size $H \times W = 7 \times 7$.  Unlike the shared module and point cloud encoder, the image encoder was pretrained on ImageNet~\cite{imagenet} and frozen. The training configuration of each network included an Adam optimizer with learning rate of 1x10$^{-6}$ and weight decay of 0.1. The learning rate was reduced by a factor of 10 every 25 epochs. 

%% file: 5_evaluation.tex
\begin{table}
  \centering
  \caption{Free-flow Speed Estimation Model Performances}
  \begin{tabular}{@{}lr@{}}
    \toprule
    \multicolumn{1}{l}{Method} & \multicolumn{1}{r}{Accuracy} \\
    \toprule
    {\em Song et al. Image Only~\cite{song2019remote}}            & 37.60\% \\
    {\em Song et al. Image + Road Features~\cite{song2019remote}} & 49.86\% \\
    \hline
    {\em Reduced PointNet++~\cite{pointnet++}}                    & 34.08\% \\
    {\em ResNet~\cite{resnet}}                                    & 42.01\% \\
    {\em RasterNet Statistics}                                    & 47.75\% \\
    {\em \textbf{RasterNet Learn}}                                & \textbf{50.47}\% \\ 
    \bottomrule
  \end{tabular}
  \label{tbl:model_performances}
\end{table}

\section{Evaluation}

We present an ablation study, a quantitative analysis compared with an existing approach on a held-out test set, and a qualitative evaluation of our best method compared with known free-flow speeds. Training, validation, and test dataset partitioning followed the methodology established by Song et al.~\cite{song2019remote}. Each model was evaluated on the set of weights chosen based on the lowest validation loss. Roads within the borders of the following Kentucky, USA counties were held-out for the test set: Bell, Lee, Ohio, Union, Woodford, Owen, Fayette, and Campbell. The validation set was constructed from 1\% of the training set samples.

\subsection{Quantitative Evaluation}

We performed an ablation study comparing different image feature extractors and the impact of point cloud features on free-flow speed estimation, shown in Table~\ref{tbl:model_performances}. Following previous work, free-flow speed estimation was evaluated using within-5mph accuracy. In this metric, predicted free-flow speed is considered correct if it is within 5mph of the true speed. 

We evaluated the performance of a full ResNet model trained only on overhead imagery in order to highlight the differences in image feature extractors compared to previous work. Specifically, we compared an Xception-style~\cite{song2019remote} architecture to our ResNet18 architecture. The first 3 blocks and the 4th block's residual sub-block were frozen, similar to the \emph{RasterNet} architectures. The smaller ResNet (12M parameters) network trained only with image features outperformed the Xception-based network (23M parameters) by 5\% average within-5mph test accuracy, suggesting it was the superior image feature extractor for this task. 

To understand the impact of augmenting point cloud features with visual features, we compared our approach to a point cloud only baseline that uses a reduced PointNet++ model as in \emph{RasterNet Learn}. Following the same strategy, the second multi-scale grouping layer was modified to extract features at raster center locations. The number of fully connected layers in the last MLP (after the multi-scale grouping layer) was reduced to two layers for faster training. The reduced PointNet++ with raster center locations had the worst performance of all of the evaluated models. While the performance is still respectable, it shows that point clouds alone do not provide the features necessary for this task.


Next, we examined the performance impact of the point cloud feature extraction strategies. We observed that the learned features (\emph{RasterNet Learn}) perform better than the structural features (\emph{RasterNet Statistics}). By combining features from both point cloud and overhead imagery, we are able to greatly improve the accuracy compared to the single modality networks. Furthermore, our \emph{RasterNet Learn} model achieves state-of-the-art performance over the previous best method, despite not using highway geometric features. In subsequent experiments,  we use the \emph{RasterNet Learn} model.


 %

\begin{figure}
  \centering
  \includegraphics[trim=70px 5px 80px 40px, clip, width=\linewidth]{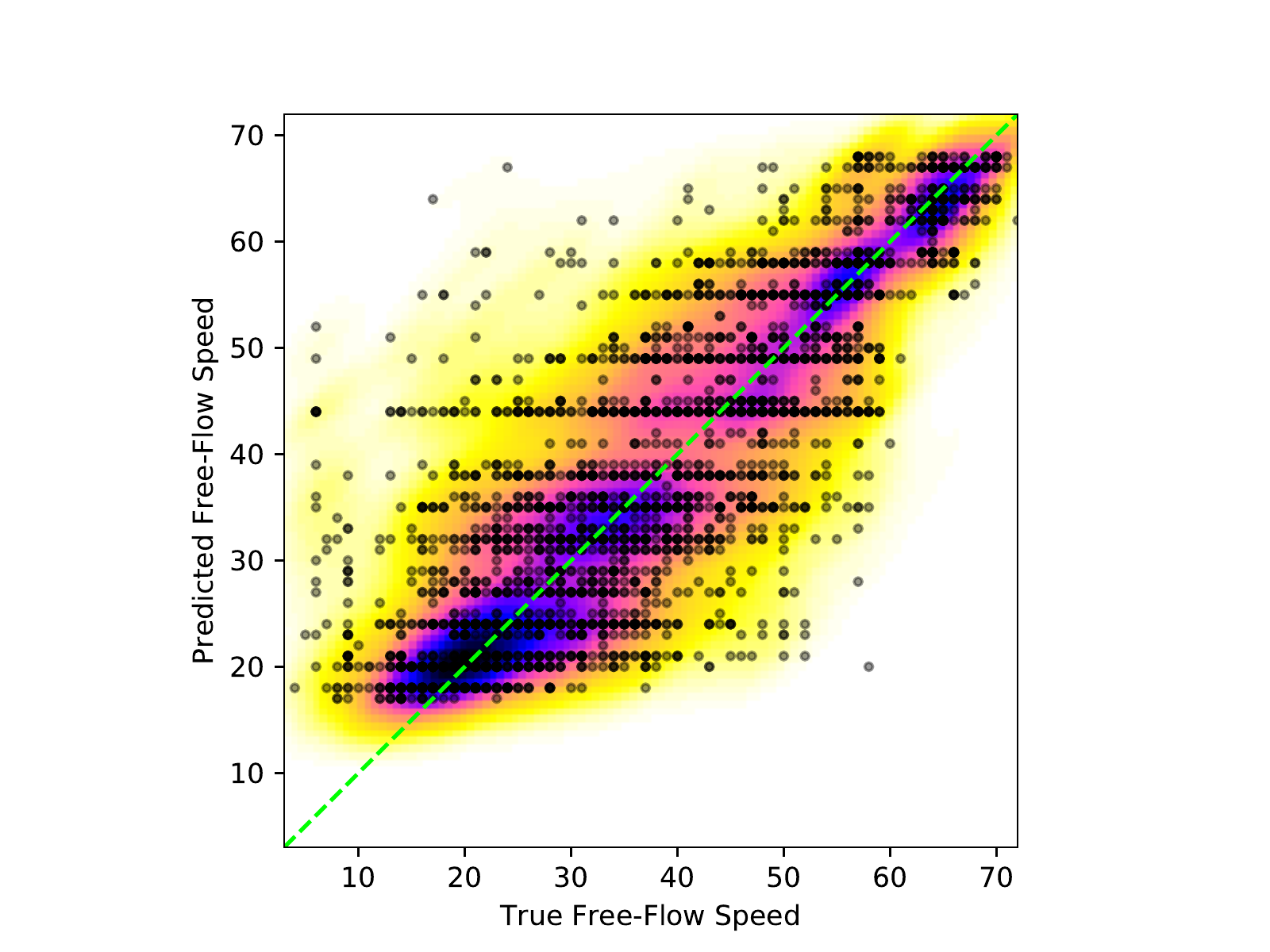} 
  \caption{Scatterplot of free-flow speed predictions on the test set from the \emph{RasterNet Learn} model compared with known speed labels. Overlayed heatmap depicts higher point density in darker color. Optimal performance should follow the green line.} 
  \label{fig:scatter_test_results}
\end{figure}

\begin{figure*}
  \centering
  \begin{subfigure}{.48\linewidth}
    \includegraphics[width=\linewidth]{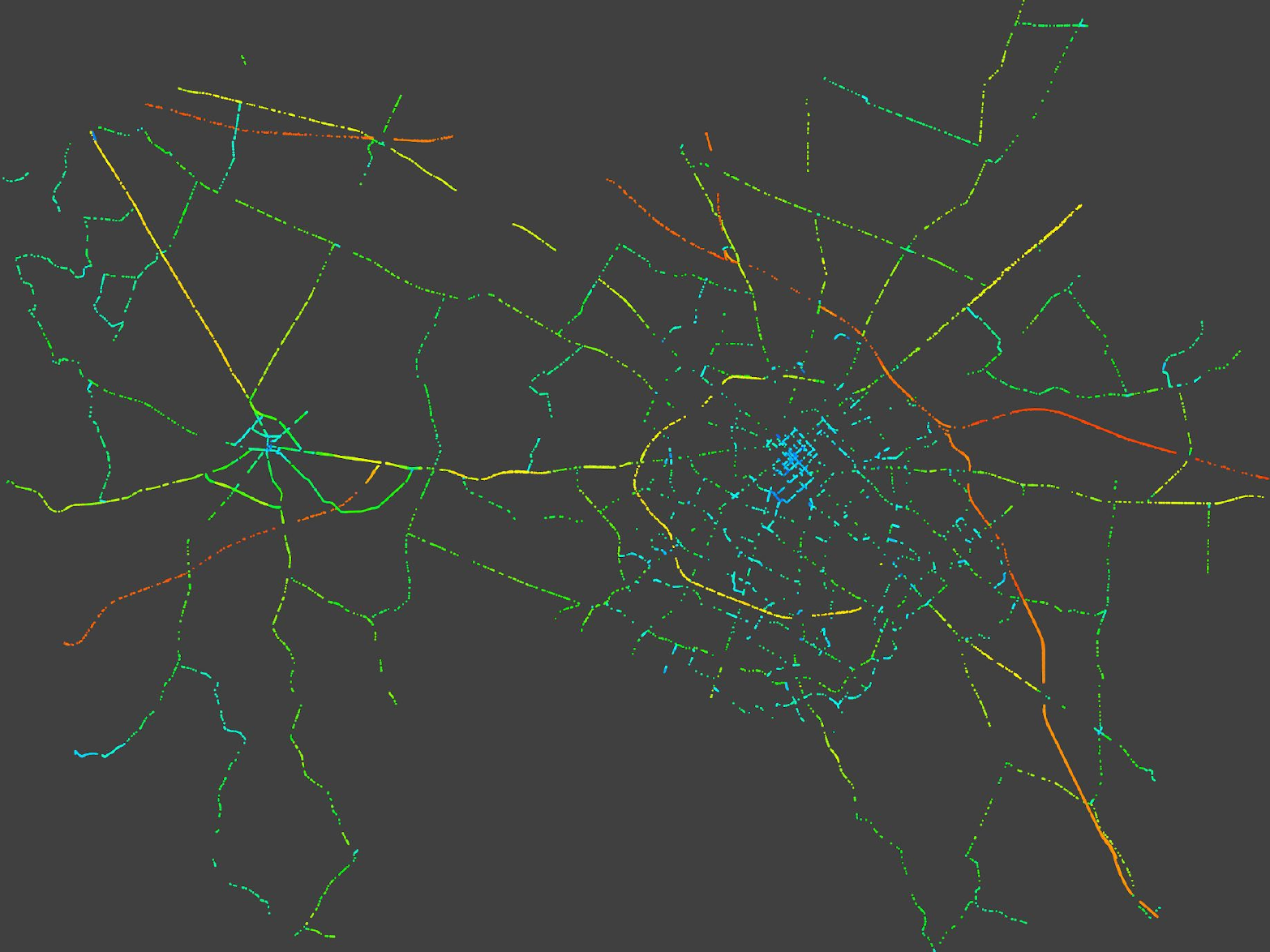}
    \caption{Fayette and Woodford Counties Ground Truth}
  \end{subfigure}
  \begin{subfigure}{.48\linewidth}
    \includegraphics[width=\linewidth]{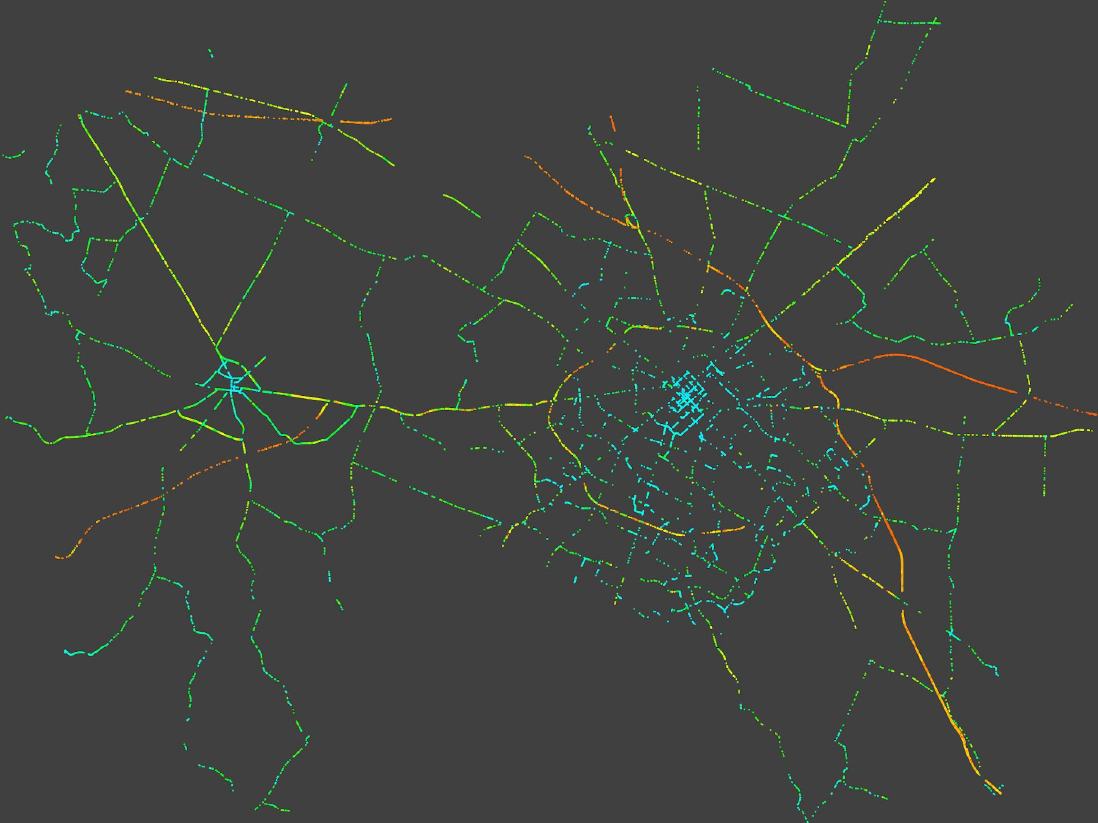}
    \caption{Fayette and Woodford Counties Predicted}
  \end{subfigure}
  
  \begin{subfigure}{.48\linewidth}
    \includegraphics[width=\linewidth]{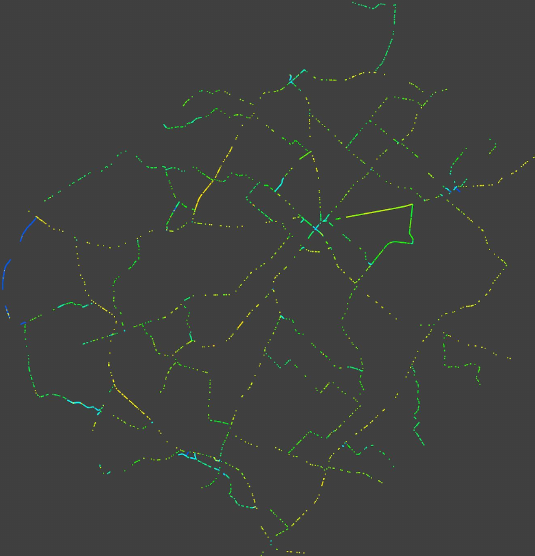}
    \caption{Union County Ground Truth}
  \end{subfigure}
  \begin{subfigure}{.48\linewidth}
    \includegraphics[width=\linewidth]{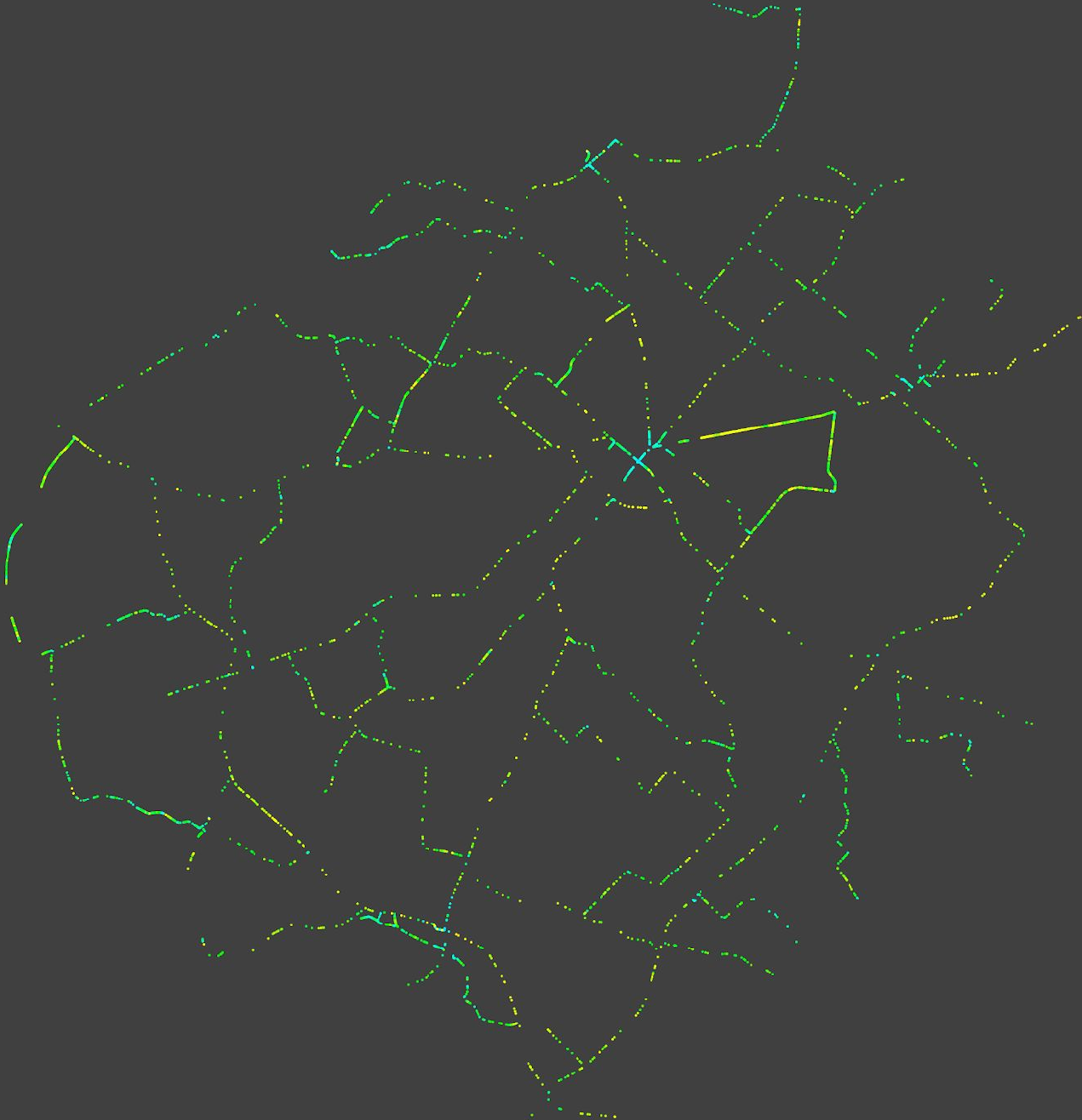}
    \caption{Union County Predicted}
  \end{subfigure}
  \includegraphics[trim=50px 0px 40px 138px, clip, width=0.5\linewidth]{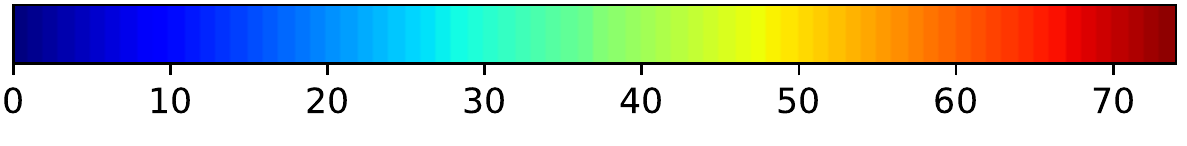}
  \caption{Ground truth and predicted speed maps for both Woodford (top left small city), Fayette (top larger city) and Union (bottom) counties in Kentucky, USA.}
  \label{fig:speed_maps}
\end{figure*}

\subsection{Qualitative Evaluation}

 Figure~\ref{fig:scatter_test_results} shows a scatter plot of the model's prediction versus ground-truth free-flow speeds on the test set. In addition, it includes a heatmap, generated using kernel density estimation, to make the joint distribution clear. Overall, the highest density of predictions (the darker colors) follow the green line, indicating a positive relationship with the true free-flow speeds. While the model had difficulties in predicting speeds accurately for roads with true free-flow speeds $<10$mph, for most other roads the model predicts speeds close to the ground-truth speed.
 

Additionally, we visualized the \emph{RasterNet Learn} model by constructing free-flow speed maps. We generated these maps with the ground truth and predicted free-flow speeds for 3 Kentucky counties from the test set: Fayette, Woodford, and Union. Since Fayette and Woodford counties are adjacent, we visualize them on the same map in Figure~\ref{fig:speed_maps} (a) and (b). Figure~\ref{fig:speed_maps} (b) suggests that the model is capable of estimating free-flow speeds on highways accurately, as shown by two major highways both being red in both maps. Unlike highways and surban areas, urban arterial road segments, as seen in the Lexington city center of Figure~\ref{fig:speed_maps} (a) and (b), are more challenging. These low speed urban arterial road segments are impacted by traffic signal timings which play a large role in regulating vehicle speeds, which are not captured in overhead imagery and LiDAR data.

The model performs well in rural counties, such as Union county in Figure~\ref{fig:speed_maps} (c) and (d), with speeds primarily ranging from 30-50mph. Note in Figure~\ref{fig:speed_maps} (c), the road segment on the far left is dark blue, indicating free-flow speeds $<20$mph. The predicted free-flow speed map in Figure~\ref{fig:speed_maps} (d) suggests that the model predicts speeds $>20$mph for said road segment. The road segment in question is a dirt road, an underrepresented road type in the training set, likely causing the poor performance in this scenario.



%% file: 6_conclusion.tex
\section{Conclusion}

We presented a novel multi-modal architecture for free-flow speed estimation, \emph{RasterNet}, that jointly processes aligned overhead images and corresponding 3D point clouds from airborne LiDAR. To support training and evaluating our methods, we introduced a large dataset of free-flow speeds, overhead imagery, and LiDAR point clouds across the state of Kentucky. We evaluated our approach on a benchmark dataset, achieving state-of-the-art results without requiring explicit highway geometric features, unlike the previous best method. Additionally, we show how our approach can be used to generate large-scale free-flow speed maps, a potentially useful tool for transportation engineering and roadway planning. Our results demonstrate that a combination of overhead imagery and 3D point clouds can replace and ultimately outperform existing approaches that rely on manually annotated input data. Our hope is that our dataset and proposed approach will inspire future work in estimating free-flow speeds from multi-modal input data.

